%% file: mid3d.tex
\documentclass{article}

\usepackage[final]{nips_2017}

\usepackage[utf8]{inputenc} 
\usepackage[T1]{fontenc}    
\usepackage{url}            
\usepackage{booktabs}       
\usepackage{amsfonts}       
\usepackage{nicefrac}       
\usepackage{microtype}      

\input{packages}
\input{macros}

\title{\model: 3D Shape Reconstruction via 2.5D Sketches}

\author{
    Jiajun Wu*\\
    MIT CSAIL\\
    \And
    Yifan Wang*\\
    ShanghaiTech University\\
    \And
    Tianfan Xue\\
    MIT CSAIL\\
    \And
    Xingyuan Sun\\
    Shanghai Jiao Tong University\\
    \And
    William T. Freeman \\
    MIT CSAIL, Google Research \\
    \And
    Joshua B. Tenenbaum \\
    MIT CSAIL
}

\begin{document}

\maketitle
\footnotetext{$*$ indicates equal contributions.} 

\input{abstract}
\input{intro}
\input{related_work}

\input{model}
\input{evaluation}

\input{discussion}

\subsection*{Acknowledgements}

We thank Shubham Tulsiani for sharing the DRC results, and Chengkai Zhang for the help on shape visualization. This work is supported by NSF \#1212849 and \#1447476, ONR MURI N00014-16-1-2007, the Center for Brain, Minds and Machines (NSF \#1231216), Toyota Research Institute, Samsung, and Shell.

\small
{
\bibliographystyle{plainnat}
\bibliography{mid3d}
}

\end{document}

%% file: packages.tex
\usepackage{color,xcolor}
\usepackage{epsfig}
\usepackage{graphicx}

\usepackage{adjustbox}
\usepackage{array}
\usepackage{booktabs}
\usepackage{colortbl}
\usepackage{float,wrapfig}
\usepackage{hhline}
\usepackage{multirow}
\usepackage{subcaption} 

\usepackage{amsmath,amsfonts,amsthm,amssymb}
\usepackage{bm}
\usepackage{nicefrac}
\usepackage{microtype}

\usepackage{changepage}
\usepackage{extramarks}
\usepackage{fancyhdr}
\usepackage{lastpage}
\usepackage{setspace}
\usepackage{soul}
\usepackage{xspace}

\usepackage[pagebackref=true,breaklinks=true,colorlinks,citecolor=gray]{hyperref}
\usepackage{url}

\usepackage{algorithm, algorithmic}
\usepackage{enumerate}
\usepackage{todonotes} 

\usepackage{titlesec}

%% file: macros.tex
\newcolumntype{L}[1]{>{\raggedright\let\newline\\\arraybackslash\hspace{0pt}}m{#1}}
\newcolumntype{C}[1]{>{\centering\let\newline\\\arraybackslash\hspace{0pt}}m{#1}}
\newcolumntype{R}[1]{>{\raggedleft\let\newline\\\arraybackslash\hspace{0pt}}m{#1}}

\newcommand{\sect}[1]{Section~\ref{#1}}

\newcommand{\fig}[1]{Figure~\ref{#1}}
\newcommand{\tbl}[1]{Table~\ref{#1}}

\newcommand{\ignorethis}[1]{}

\makeatletter
\DeclareRobustCommand\onedot{\futurelet\@let@token\@onedot}
\def\@onedot{\ifx\@let@token.\else.\null\fi\xspace}

\def\ie{\emph{i.e}\onedot} 
 
\def\etc{\emph{etc}\onedot} \def\vs{\emph{vs}\onedot}

\makeatother

\definecolor{MyDarkBlue}{rgb}{0,0.08,1}
\definecolor{MyDarkGreen}{rgb}{0.02,0.6,0.02}
\definecolor{MyDarkRed}{rgb}{0.8,0.02,0.02}
\definecolor{MyDarkOrange}{rgb}{0.40,0.2,0.02}
\definecolor{MyPurple}{RGB}{111,0,255}
\definecolor{MyRed}{rgb}{1.0,0.0,0.0}
\definecolor{MyGold}{rgb}{0.75,0.6,0.12}
\definecolor{MyDarkgray}{rgb}{0.66, 0.66, 0.66}

\newcommand{\myparagraph}[1]{\vspace{-3pt}\paragraph{#1}}

\newcommand{\model}{MarrNet\xspace}

%% file: abstract.tex
\vspace{-10pt}
\begin{abstract}

3D object reconstruction from a single image is a highly under-determined problem, requiring strong prior knowledge of plausible 3D shapes. This introduces challenges for learning-based approaches, as 3D object annotations are scarce in real images. Previous work chose to train on synthetic data with ground truth 3D information, but suffered from domain adaptation when tested on real data.

In this work, we propose \model, an end-to-end trainable model that sequentially estimates 2.5D sketches and 3D object shape. Our disentangled, two-step formulation has three advantages. First, compared to full 3D shape, 2.5D sketches are much easier to be recovered from a 2D image; models that recover 2.5D sketches are also more likely to transfer from synthetic to real data. Second, for 3D reconstruction from 2.5D sketches, systems can learn purely from synthetic data. This is because we can easily render realistic 2.5D sketches without modeling object appearance variations in real images, including lighting, texture, \etc. This further relieves the domain adaptation problem. Third, we derive differentiable projective functions from 3D shape to 2.5D sketches; the framework is therefore end-to-end trainable on real images, requiring no human annotations. Our model achieves state-of-the-art performance on 3D shape reconstruction. 

\end{abstract}

%% file: intro.tex
\section{Introduction}
\label{sec:intro}

Humans quickly recognize 3D shapes from a single image. \fig{fig:teaser}a shows a number of images of chairs; despite their drastic difference in object texture, material, environment lighting, and background, humans easily recognize they have very similar 3D shapes. What is the most essential information that makes this happen?

Researchers in human perception argued that our 3D perception could rely on recovering 2.5D sketches~\citep{Marr1982}, which include intrinsic images~\citep{barrow,tappen2003recovering} like depth and surface normal maps (\fig{fig:teaser}b). Intrinsic images disentangle object appearance variations in texture, albedo, lighting, \etc, with its shape, which retains all information from the observed image for 3D reconstruction. Humans further combine 2.5D sketches and a shape prior learned from past experience to reconstruct a full 3D shape (\fig{fig:teaser}c). In the field of computer vision, there have also been abundant works exploiting the idea for reconstruction 3D shapes of faces~\citep{kemelmacher20113d}, objects~\citep{tappen2003recovering}, and scenes~\citep{Hoiem2005b,saxena2009make3d}.

Recently, researchers attempted to tackle the problem of single image 3D reconstruction with deep learning. These approaches usually regress a 3D shape from a single RGB image directly~\citep{tulsiani2017multi,Choy2016,Wu2016}. In contrast, we propose a two-step while end-to-end trainable pipeline, sequentially recovering 2.5D sketches (depth and normal maps) and a 3D shape. We use an encoder-decoder structure for each component of the framework, and also enforce the reprojection consistency between the estimated sketch and the 3D shape. We name it MarrNet, for its close resemblance to David Marr's theory of perception~\citep{Marr1982}.

Our approach offers several unique advantages. First, the use of 2.5D sketches releases the burden on domain transfer. As single image 3D reconstruction is a highly under-constrained problem, strong prior knowledge of object shapes is needed. This poses challenges to learning-based methods, as accurate 3D object annotations in real images are rare. Most previous methods turned to training purely on synthetic data~\citep{tulsiani2017multi,Choy2016,Girdhar2016}. However, these approaches often suffer from the domain adaption issue due to imperfect rendering. Learning 2.5D sketches from images, in comparison, is much easier and more robust to transfer from synthetic to real images, as shown in \sect{sec:eval}.

Further, as our second step recovers 3D shape from 2.5D sketches --- an abstraction of the raw input image, it can be trained purely relying on synthetic data. Though rendering diverse realistic images is challenging, it is straightforward to obtain almost perfect object surface normals and depths from a graphics engine. This further relieves the domain adaptation issue. 

We also enforce differentiable constraints between 2.5D sketches and 3D shape, making our system end-to-end trainable, even on real images without any annotations. Given a set of unlabeled images, our algorithm, pre-trained on synthetic data, can infer the 2.5D sketches of objects in the image, and use it to refine its estimation of objects' 3D shape. This self-supervised feature enhances its performance on images from different domains.

We evaluate our framework on both synthetic images of objects from ShapeNet~\citep{Chang2015}, and real images from the PASCAL 3D+ dataset~\citep{Xiang2014}. We demonstrate that our framework performs well on 3D shape reconstruction, both qualitatively and quantitatively. 

Our contributions are three-fold: inspired by visual cognition theory, we propose a two-step, disentangled formulation for single image 3D reconstruction via 2.5D sketches; 
we develop a novel, end-to-end trainable model with a differentiable projection layer that ensures consistency between 3D shape and mid-level representations; 
we demonstrate its effectiveness on 2.5D sketch transfer and 3D shape reconstruction on both synthetic and real data. 

\input{figText_teaser}

%% file: figText_teaser.tex
\begin{figure}[t]
    \centering
    \includegraphics[width=\linewidth]{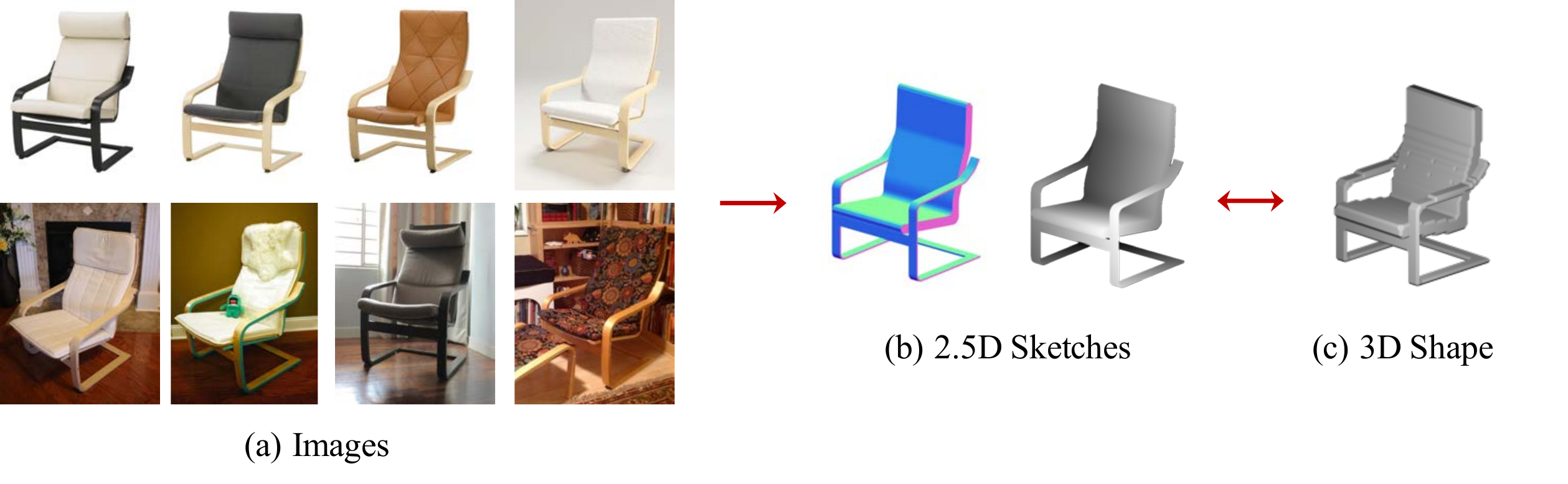}
    \caption{Objects in real images (a) are subject to appearance variations regarding color, texture, lighting, material, background, \etc. Despite this, their 2.5D sketches like surface normal and depth maps remain constant (b). The 2.5D sketches can be seen as an abstraction of the image, retaining all information about the 3D shape of the object inside. We combine the sketches with learned shape priors to reconstruct the full 3D shape (c).}
    \vspace{-8pt}
    \label{fig:teaser}
\end{figure}

%% file: related_work.tex
\section{Related Work}

\paragraph{2.5D Sketch Recovery} Estimating 2.5D sketches has been a long-standing problem in computer vision. In the past, researchers have explored recovering 2.5D shape from shading, texture, or color images~\citep{Horn1989,zhang1999shape,tappen2003recovering,Barron2015,Weiss2001b,Bell2014}. With the development of depth sensors~\citep{Izadi2011} and larger scale RGB-D datasets~\citep{Silberman2012,song2016semantic,mccormac2016scenenet}, there have also been papers on estimating depth~\citep{chen2016single,Eigen2015}, surface normals~\citep{Bansal2016,Wang2015b}, and other intrinsic images~\citep{Shi2017Learning,rin} with deep networks. Our method employs 2.5D estimation as a component, but targets reconstructing full 3D shape of an object. 

\myparagraph{Single Image 3D Reconstruction} The problem of recovering object shape from a single image is challenging, as it requires both powerful recognition systems and prior shape knowledge. With the development of large-scale shape repository like ShapeNet~\citep{Chang2015}, researchers developed models encoding shape prior for this task~\citep{Girdhar2016,Choy2016,tulsiani2017multi,Wu2016,Kar2015,kanazawa2016warpnet,soltani2017synthesizing}, with extension to scenes~\citep{song2016semantic}. These methods typically regress a voxelized 3D shape directly from an input image, and rely on synthetic data or 2D masks for training. In comparison, our formulation tackles domain difference better, as it can be end-to-end fine-tuned on images without any annotations.

\myparagraph{2D-3D Consistency} It is intuitive and practically helpful to constrain the reconstructed 3D shape to be consistent with 2D observations. Researchers have explored this idea for decades~\citep{Lowe1987}. This idea is also widely used in 3D shape completion from depths or silhouettes~\citep{Firman2016Structured,Rock2015Completing,Dai2017Shape}. Recently, a few papers discussed enforcing differentiable 2D-3D constraints between shape and silhouettes, enabling joint training of deep networks for 3D reconstruction~\citep{Wu2016b,Yan2016,Rezende2016,tulsiani2017multi}. In our paper, we exploit this idea to develop differentiable constraints on the consistency between various 2.5D sketches and 3D shape.

%% file: model.tex
\section{Approach}
\label{sec:model}

\input{figText_model}
\label{sec:step1}

To recover the 3D structure from a single view RGB image, our MarrNet contains three parts: first, a 2.5D sketch estimator, which predicts the depth, surface normal, and silhouette images of the object (\fig{fig:model}a); second, a 3D shape estimator, which infers 3D object shape using a voxel representation (\fig{fig:model}b); third, a reprojection consistency function, enforcing the alignment between the estimated 3D structure and inferred 2.5D sketches (\fig{fig:model}c).

\subsection{2.5D Sketch Estimation}
The first component of our network (\fig{fig:model}a) takes a 2D RGB image as input, and predicts its 2.5D sketch: surface normal, depth, and silhouette. The goal of the 2.5D sketch estimation step is to distill intrinsic object properties from input images, while discarding properties that are non-essential for the task of 3D reconstruction, such as object texture and lighting.

We use an encoder-decoder network architecture for 2.5D sketch estimation. Our encoder is a ResNet-18~\citep{He2015}, encoding a 256$\times$256 RGB image into 512 feature maps of size 8$\times$8. The decoder contains four sets of 5$\times$5 fully convolutional and ReLU layers, followed by four sets of 1$\times$1 convolutional and ReLU layers. It outputs the corresponding depth, surface normal, and silhouette images, also at the resolution of 256$\times$256. 

\subsection{3D Shape Estimation}
\label{sec:step2}

The second part of our framework (\fig{fig:model}b) infers 3D object shape from estimated 2.5D sketches. Here, the network focuses on learning the shape prior that explains input well. As it takes only surface normal and depth images as input, it can be trained on synthetic data, without suffering from the domain adaption problem: it is straightforward to render nearly perfect 2.5D sketches, but much harder to render realistic images. 

The network architecture is inspired by the TL network~\citep{Girdhar2016}, and the 3D-VAE-GAN~\citep{Wu2016}, again with an encoding-decoding style. It takes a normal image and a depth image as input (both masked by the estimated silhouette), maps them to a 200-dim vector via five sets of convolutional, ReLU, and pooling layers,  followed by two fully connected layers. The detailed encoder structure can be found in \cite{Girdhar2016}. The vector then goes through a decoder, which consists of five fully convolutional and ReLU layers to output a 128$\times$128$\times$128 voxel-based reconstruction of the input. The detailed encoder structure can be found in \cite{Wu2016}.

\subsection{Reprojection Consistency}
\label{sec:reprojection}

\input{figText_projection}

There have been works attempting to enforce the consistency between estimated 3D shape and 2D representations in a neural network~\citep{Yan2016,Rezende2016,Wu2016b,tulsiani2017multi}. Here, we explore novel ways to include a reprojection consistency loss between the predicted 3D shape and the estimated 2.5D sketch, consisting of a depth reprojection loss and a surface normal reprojection loss. 

We use $v_{x, y, z}$ to represent the value at position $(x, y, z)$ in a 3D voxel grid, assuming that $v_{x, y, z}\in[0,1]$, $\forall x, y, z$. We use $d_{x, y}$ to denote the estimated depth at position $(x, y)$, and $n_{x, y} = (n_a, n_b, n_c)$ to denote the estimated surface normal. We assume orthographic projection in this work. 

\myparagraph{Depths}
The projected depth loss tries to guarantee that the voxel with depth $v_{x, y, d_{x, y}}$ should be 1, and all voxels in front of it should be $0$. This ensures the estimated 3D shape matches the estimated depth values.

As illustrated in \fig{fig:projection}a, we define projected depth loss as follows:
\begin{equation}
		L_\text{depth}(x, y, z) = 
		\begin{cases}
			v^2_{x, y, z}, & z < d_{x, y} \\
			(1 - v_{x, y, z})^2, & z = d_{x, y} \\
			0, & z > d_{x, y} \\
		\end{cases}.
\end{equation}
The gradients are
\begin{equation}
		\frac{\partial{L_\text{depth}(x, y, z)}}{\partial{v_{x, y, z}}} =
		\begin{cases}
			2 v_{x, y, z}, & z < d_{x, y} \\
			2 (v_{x, y, z} - 1), & z = d_{x, y} \\
			0, & z > d_{x, y} \\
		\end{cases}.
\end{equation}

When $d_{x,y}=\infty$, our depth criterion reduces to a special case --- the silhouette criterion. As shown in \fig{fig:projection}b, for a line that has no intersection with the shape, all voxels in it should be 0.
	
\myparagraph{Surface normals}
As vectors $n_x = (0, -n_c, n_b)$ and $n_y = (-n_c, 0, n_a)$ are orthogonal to the normal vector $n_{x, y} = (n_a, n_b, n_c)$, we can normalize them to obtain two vectors, $n_x' = (0, -1, n_b/n_c)$ and $n_y' = (-1, 0, n_a/n_c)$, both on the estimated surface plane at $(x, y, z)$. The projected surface normal loss tries to guarantee that the voxels at $(x, y, z) \pm n_x'$ and $(x, y, z) \pm n_{+y}$ should be 1 to match the estimated surface normals. These constraints only apply when the target voxels are inside the estimated silhouette.

As shown in \fig{fig:projection}c, let $z = d_{x, y}$, the projected surface normal loss is defined as
\begin{align}
    L_\text{normal}(x, y, z) = \;& 
			\left(1 - v_{x, y - 1, z + \frac{n_b}{n_c}}\right) ^ 2 +
			\left(1 - v_{x, y + 1, z - \frac{n_b}{n_c}}\right) ^ 2 +  \nonumber \\
			&\left(1 - v_{x - 1, y, z + \frac{n_a}{n_c}}\right) ^ 2 +
			\left(1 - v_{x + 1, y, z - \frac{n_a}{n_c}}\right) ^ 2.
\end{align}
Then the gradients along the $x$ direction are
\begin{equation}
	\frac{\partial{L_\text{normal}(x, y, z)}}{\partial{v_{x - 1, y, z + \frac{n_a}{n_c}}}} = 2 \left(v_{x - 1, y, z + \frac{n_a}{n_c}} - 1\right)
\quad\text{and}\quad
	\frac{\partial{L_\text{normal}(x, y, z)}}{\partial{v_{x + 1, y, z - \frac{n_a}{n_c}}}} = 2 \left(v_{x + 1, y, z - \frac{n_a}{n_c}} - 1\right).
\end{equation}
The gradients along the $y$ direction are similar.

\subsection{Training paradigm}
\label{sec:train}

We employ a two-step training paradigm. We first train the 2.5D sketch estimation and the 3D shape estimation components separately on synthetic images; we then fine-tune the network on real images.

For pre-training, we use synthetic images of ShapeNet objects. The 2.5D sketch estimator is trained using the ground truth surface normal, depth, and silhouette images with a L2 loss. The 3D interpreter is trained using ground truth voxels and a cross-entropy loss. Please see \sect{sec:shapenet} for details on data preparation. 

The reprojection consistency loss is used to fine-tune the 3D estimation component of our model on real images, using the predicted normal, depth, and silhouette. We observe that a straightforward implementation leads to shapes that explain 2.5D sketches well, but with unrealistic appearance. This is because the 3D estimation module overfits the images without preserving the learned 3D shape prior. See \fig{fig:ablation} for examples, and \sect{sec:pascal} for more details.

We therefore choose to fix the decoder of the 3D estimator and only fine-tune the encoder. During testing, our method can be self-supervised, \ie, we can fine-tune even on a single image without any annotations. In practice, we fine-tune our model separately on each image for 40 iterations. For each test image, fine-tuning takes up to 10 seconds on a modern GPU; without fine-tuning, testing time is around 100 milliseconds. We use SGD for optimization with a batch size of 4, a learning rate of 0.001, and a momentum of 0.9. We implemented our framework in Torch7~\citep{Collobert2011}. 

\input{figText_shapenet}
\input{figText_ablation}

%% file: figText_model.tex
\begin{figure}[t]
    \centering
    \includegraphics[width=\linewidth]{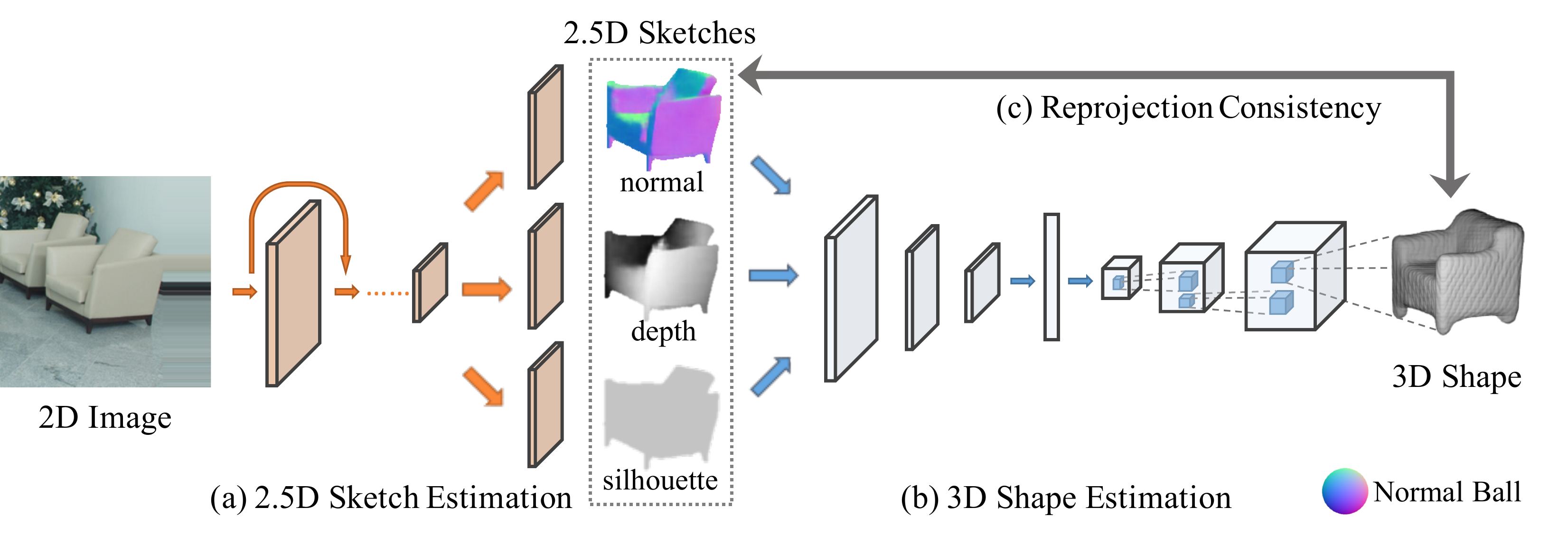}
    \caption{Our model (\model) has three major components: (a) 2.5D sketch estimation, (b) 3D shape estimation, and (c) a loss function for reprojection consistency. \model first recovers object normal, depth, and silhouette images from an RGB image. It then regresses the 3D shape from the 2.5D sketches. In both steps, it uses an encoding-decoding network. It finally employs a reprojection consistency loss to ensure the estimated 3D shape aligns with the 2.5D sketches. The entire framework can be trained end-to-end.} 
    \vspace{-5pt}
    \label{fig:model}
\end{figure}

%% file: figText_projection.tex
\begin{figure}[t]
    \centering
    \includegraphics[width=\linewidth]{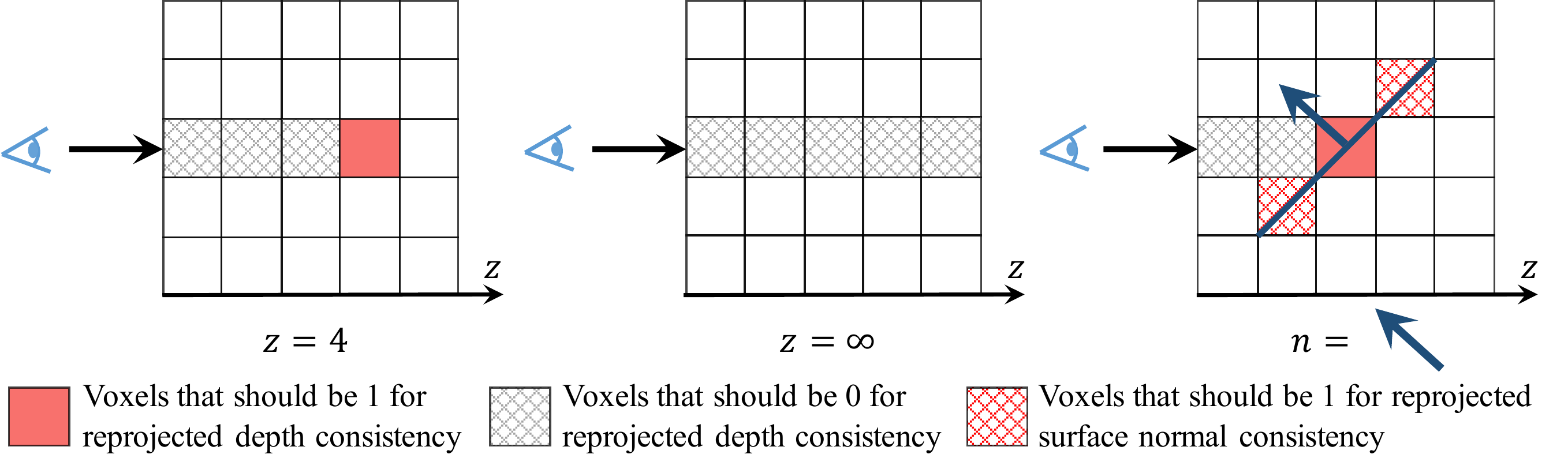}
    \caption{Reprojection consistency between 2.5D sketches and 3D shape. Left and middle: the criteria for depths and silhouettes; right: the criterion for surface normals. See \sect{sec:reprojection} for details.}
    \vspace{-5pt}
    \label{fig:projection}
\end{figure}

%% file: figText_shapenet.tex
\begin{figure}[t]
    \centering
    \includegraphics[width=\linewidth]{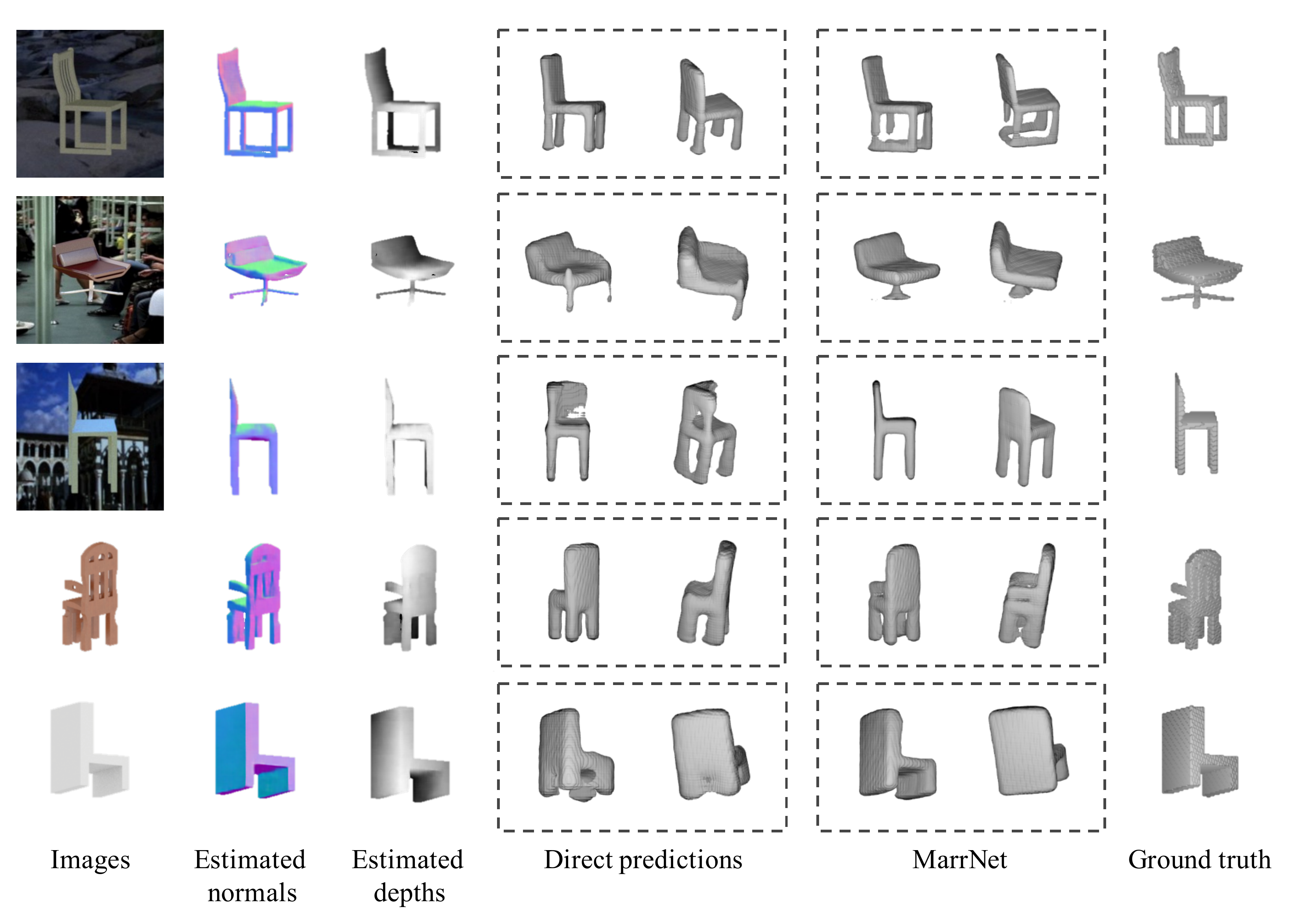}
    \vspace{-10pt}
    \caption{Results on rendered images of ShapeNet objects~\citep{Chang2015}. From left to right: input, estimated normal map, estimated depth map, our prediction, a baseline algorithm that predicts 3D shape directly from RGB input without modeling 2.5D sketch, and ground truth. Both normal and depth maps are masked by predicted silhouettes. Our method is able to recover shapes with smoother surfaces and finer details.}
    \vspace{-5pt}
    \label{fig:shapenet}
\end{figure}

%% file: figText_ablation.tex
\begin{figure}[t]
    \centering
    \includegraphics[width=\linewidth]{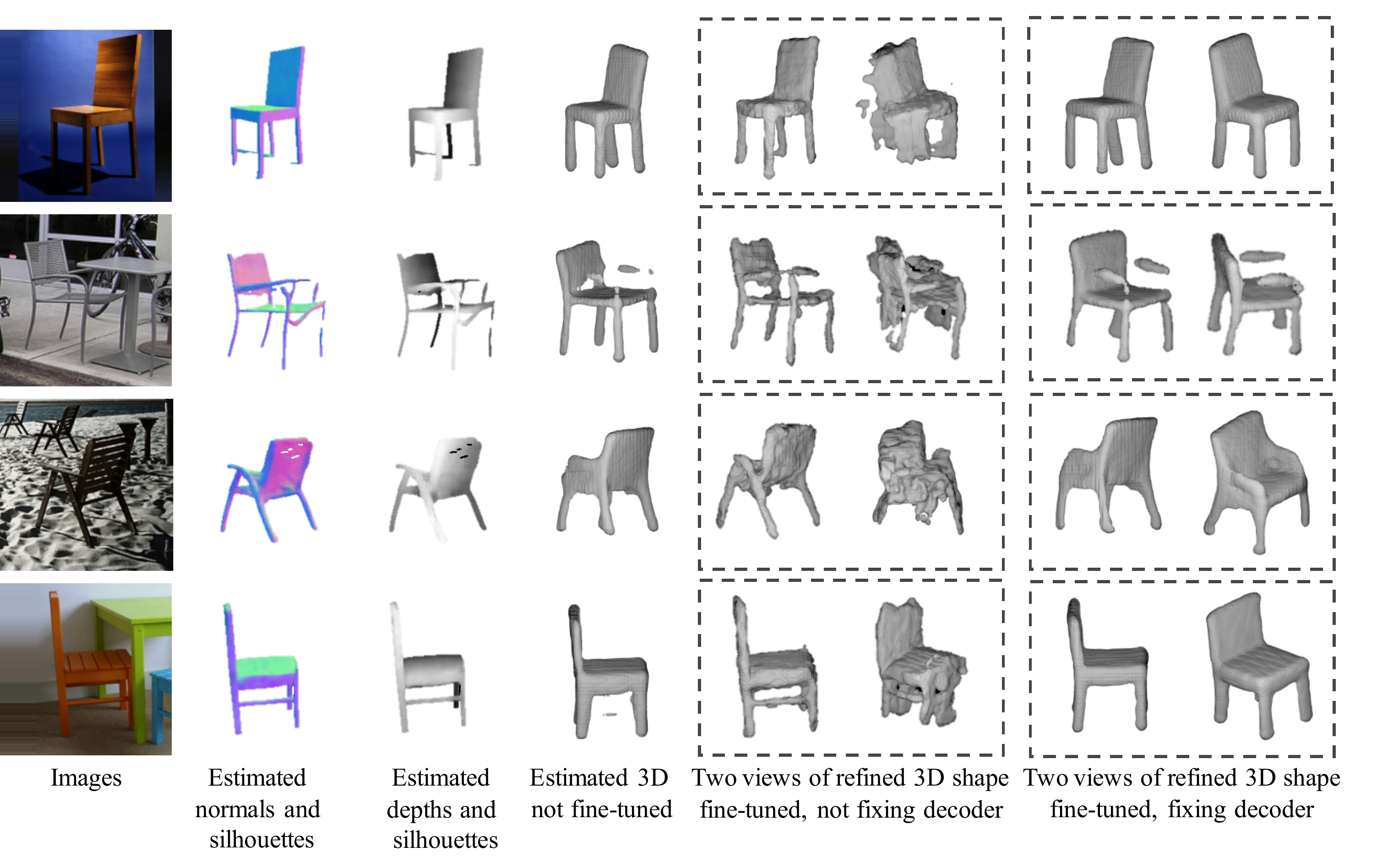}
    \caption{We present an ablation study, where we compare variants of our models. From left to right: input, estimated normal, estimated depth, 3D prediction before fine-tuning, two views of the 3D prediction after fine-tuning without fixing decoder, and two views of the 3D prediction after fine-tuning with the decoder fixed. When the decoder is not fixed, the model explains the 2.5D sketch well, but fails to preserve the learned shape prior. Fine-tuning with a fixed decoder resolves the issue.}
    \vspace{-5pt}
    \label{fig:ablation}
\end{figure}

%% file: evaluation.tex
\section{Evaluation}
\label{sec:eval}

\input{figText_pascal}
\input{figText_amt}
\input{figText_ikea}
\input{figText_multi}
\input{figText_multiclass}

In this section, we present both qualitative and quantitative results on single image 3D reconstruction using variants of our framework. We evaluate our entire framework on both synthetic and real-life images on three datasets. 

\subsection{3D Reconstruction on ShapeNet}
\label{sec:shapenet}

\paragraph{Data} 
We start with experiments on synthesized images of ShapeNet chairs~\citep{Chang2015}. We put objects in front of random backgrounds from the SUN database~\citep{Xiao2010}, and render the corresponding RGB, depth, surface normal, and silhouette images. We use a physics-based renderer, Mitsuba~\citep{Mitsuba}, to obtain more realistic images. For each of the 6,778 ShapeNet chairs, we render 20 images of random viewpoints.

\myparagraph{Methods}
We follow the training paradigm described in \sect{sec:train}, but without the final fine-tuning stage, as ground truth 3D shapes are available on this synthetic dataset. Specifically, the 2.5D sketch estimator is trained using ground truth depth, normal and silhouette images and a L2 reconstruction loss. The 3D shape estimation module takes in the masked ground truth depth and normal images as input, and predicts 3D voxels of size 128$\times$128$\times$128 with a binary cross entropy loss. 

We compare MarrNet with a baseline that predicts 3D shape directly from an RGB image, without modeling 2.5D sketches. The baseline employs the same architecture as our 3D shape estimator (\sect{sec:step2}). We show qualitative results in \fig{fig:shapenet}. Our estimated surface normal and depth images abstract out non-essential information like textures and lighting in the RGB image, while preserving intrinsic information about object shape. Compared with the direct prediction baseline, our model outputs objects with more details and smoother surfaces. For quantitative evaluation, previous works usually compute the Intersection-over-Union (IoU)~\citep{tulsiani2017multi,Choy2016}. Our full model achieves a higher IoU (0.57) than the direct prediction baseline (0.52).

\subsection{3D Reconstruction on Pascal 3D+}
\label{sec:pascal}

\paragraph{Data} 
PASCAL 3D+ dataset~\citep{Xiang2014} provides (rough) 3D models for objects in real-life images. Here, we use the same test set of PASCAL 3D+ with earlier works~\citep{tulsiani2017multi}.

\myparagraph{Methods}
We follow the paradigm described in \sect{sec:train}: we first train each module separately on the ShapeNet dataset, and then fine-tune them on the PASCAL 3D+ dataset. Unlike previous works~\citep{tulsiani2017multi}, our model requires no silhouettes as input during fine-tuning; it instead estimates silhouette jointly. 

As an ablation study, we compare three variants of our model: first, the model trained using ShapeNet data only, without fine-tuning; second, the fine-tuned model whose decoder is not fixed during fine-tuning; and third, the full model whose decoder is fixed during fine-tuning. We also compare with the state-of-the-art method (DRC)~\citep{tulsiani2017multi}, and the provided ground truth shapes.

\myparagraph{Results}
The results of our ablation study are shown in \fig{fig:ablation}. The model trained on synthetic data provides a reasonable shape estimate. If we fine-tune the model on Pascal 3D+ without fixing the decoder, the output voxels explain the 2.5D sketch data well, but fail to preserve the learned shape prior, leading to impossible shapes from certain views. Our final model, fine-tuned with the decoder fixed, keeps the shape prior and provides more details of the shape. 

We show more results in \fig{fig:pascal}, where we compare with the state-of-the-art (DRC)~\citep{tulsiani2017multi}, and the provided ground truth shapes. Quantitatively, our algorithm achieves a higher IoU over these methods (MarrNet 0.39 \vs DRC 0.34). However, we find the IoU metric sub-optimal for three reasons. First, measuring 3D shape similarity is a challenging yet unsolved problem, and IoU prefers models that predict mean shapes consistently, with no emphasize on details. Second, as object shape can only be reconstructed up to scale from a single image, it requires searching over all possible scales during the computation of IoU, making it less efficient. Third, as discussed in~\cite{tulsiani2017multi}, PASCAL 3D+ has only rough 3D annotations (10 CAD chair models for all images). Computing IoU with these shapes would thus not be the most informative evaluation metric. 

We instead conduct human studies, where we show users the input image and two reconstructions, and ask them to choose the one that looks closer to the shape in the image. We show each test image to 10 human subjects. As shown in \tbl{tbl:amt}, our reconstruction is preferred 74\% of the time to DRC, and 42\% of the time to ground truth, showing a clear advantage. 

We present some failure cases in \fig{fig:fail}. Our algorithm does not perform well on recovering complex, thin structure, and sometimes fails when the estimated mask is very inaccurate. Also, while DRC may benefit from multi-view supervision, we have only evaluated \model given a single view of the shape, though adapting our formulation to multi-view data should be straightforward. 

\subsection{3D Reconstruction on IKEA}
\paragraph{Data} 
The IKEA dataset~\citep{Lim2013} contains images of IKEA furniture, along with accurate 3D shape and pose annotations. These images are challenging, as objects are often heavily occluded or cropped. We also evaluate our model on the IKEA dataset. 

\myparagraph{Results}
We show qualitative results in \fig{fig:ikea}, where we compare with estimations by 3D-VAE-GAN~\citep{Wu2016} and the ground truth. As shown in the figure, our model can deal with mild occlusions in real life scenarios. We also conduct human studies on the IKEA dataset. Results show that 61\% of the subjects prefer our reconstructions to those of 3D-VAE-GAN. 

\subsection{Extensions}

We also apply our framework on cars and airplanes. We use the same setup as that for chairs. As shown in \fig{fig:multi}, shape details like the horizontal stabilizer and rear-view mirrors are recovered by our model. We further train \model jointly on all three object categories, and show results in \fig{fig:multiclass}. Our model successfully recovers shapes of different categories.

%% file: figText_pascal.tex
\begin{figure}[t]   
  \includegraphics[width=\linewidth]{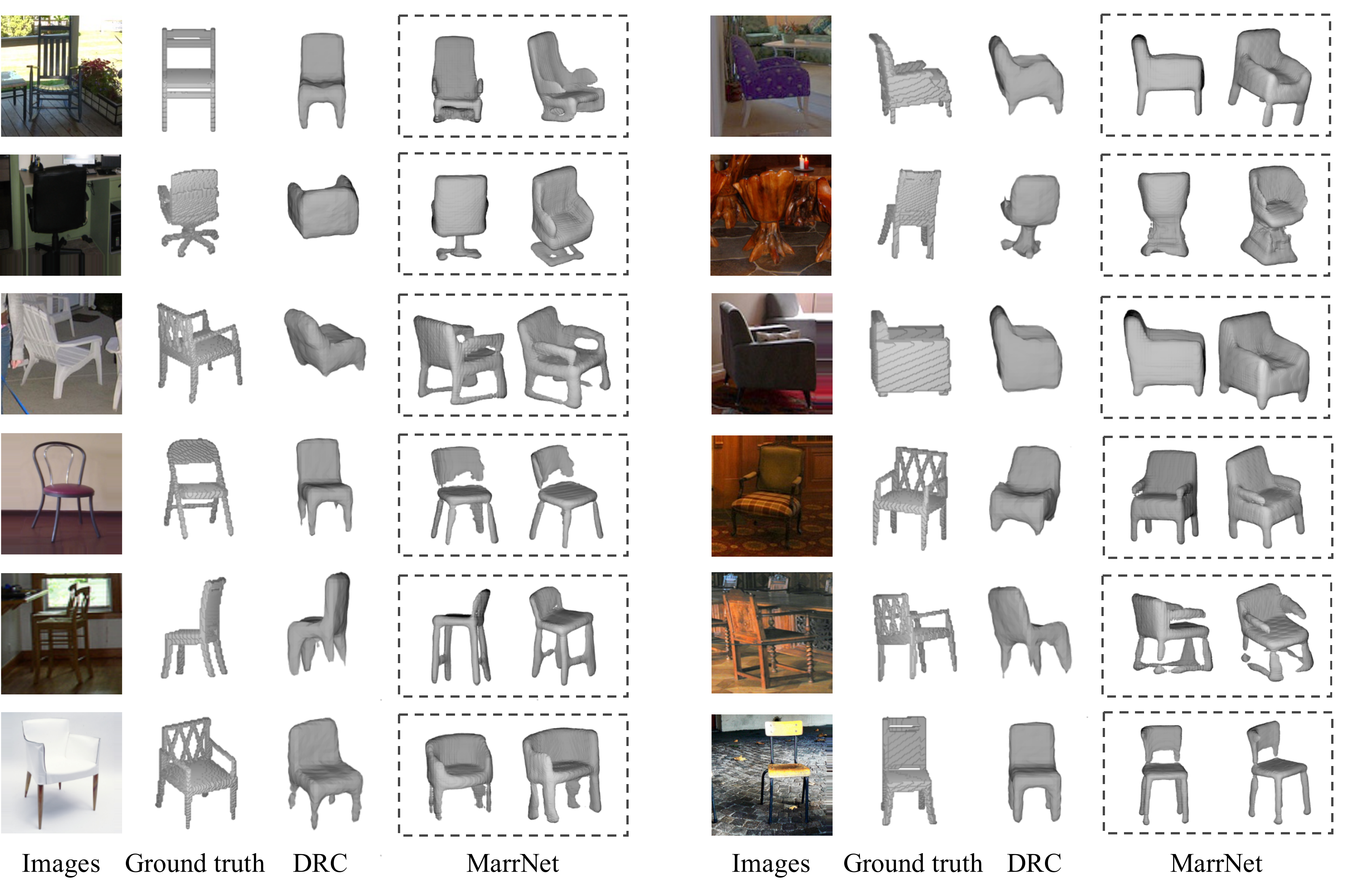}
    \caption{3D reconstructions of chairs on the Pascal 3D+~\citep{Xiang2014} dataset. From left to right: input, the ground truth shape from the dataset, 3D estimation by DRC~\citep{tulsiani2017multi}, and two views of \model predictions. Our model recovers more accurate 3D shapes. }
    \vspace{-5pt}
    \label{fig:pascal}
\end{figure}

%% file: figText_amt.tex
\begin{figure}[t]
\begin{minipage}[c]{0.48\linewidth}
 	\centering
    \begin{tabular}{lccc}
    \toprule
     & DRC & \model & GT \\
    \midrule
    DRC & 50 & 26 & 17 \\
    \model & 74 & 50 & 42 \\
    Ground truth & 83 & 58 & 50\\
    \bottomrule
    \end{tabular}
    \vspace{10pt}
    \captionof{table}{Human preferences on chairs in PASCAL 3D+~\citep{Xiang2014}. We compare \model with the state-of-the-art (DRC)~\citep{tulsiani2017multi}, and the ground truth provided by the dataset. Each number shows the percentage of humans prefer the left method to the top one. \model is preferred 74\% of the time to DRC, and 42\% of the time to ground truth.}
    \label{tbl:amt}
\end{minipage}
\hfill
\begin{minipage}[c]{0.48\linewidth}
    \centering
    \includegraphics[width=\linewidth]{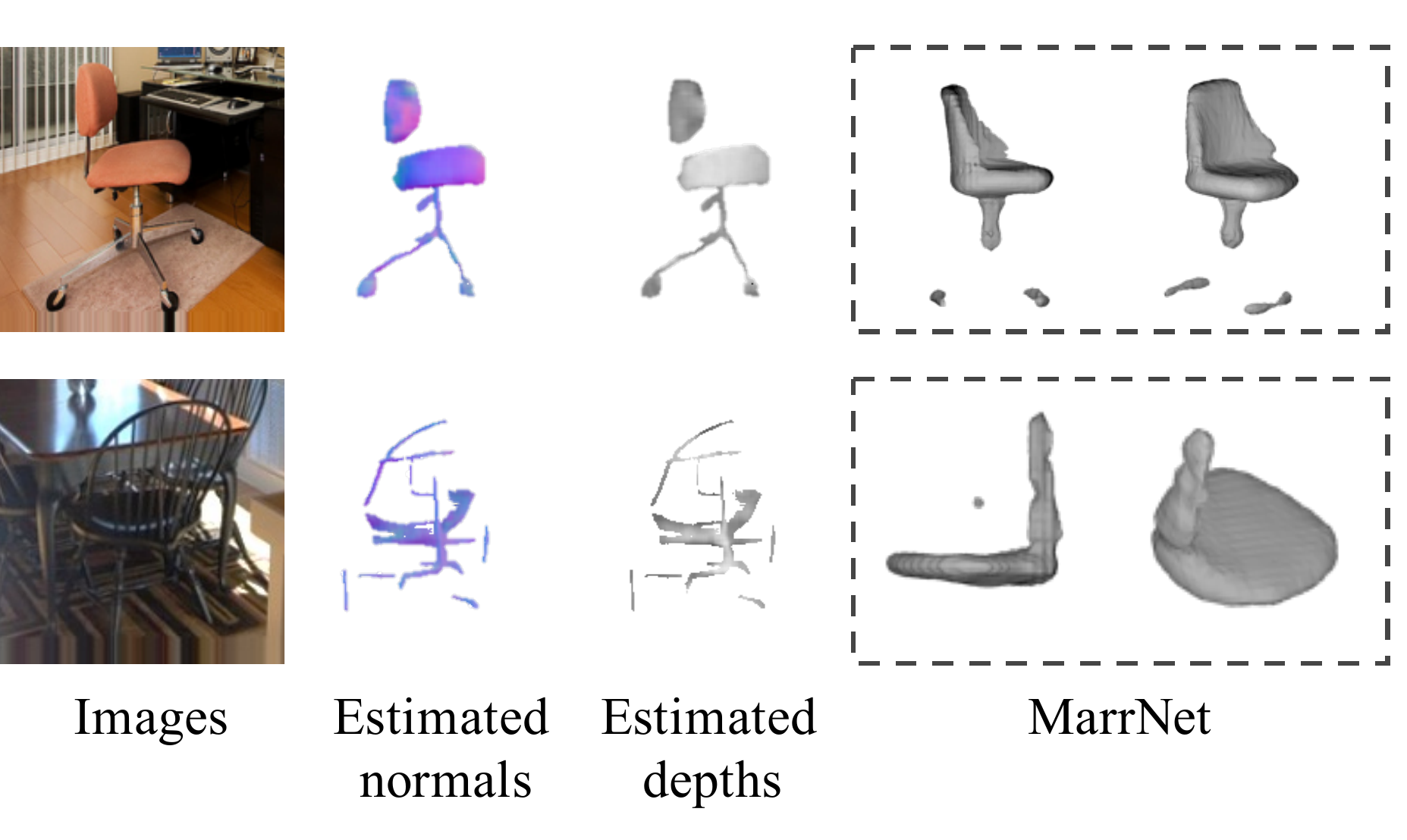}
    \captionof{figure}{Failure cases on Pascal 3D+. Our algorithm does not perform well on recovering complex, thin structure, and sometimes fails when the estimated mask is very inaccurate.}
    \label{fig:fail}
\end{minipage}
\vspace{-5pt}
\end{figure}

%% file: figText_ikea.tex
\begin{figure}[t]
    \centering
    \includegraphics[width=\linewidth]{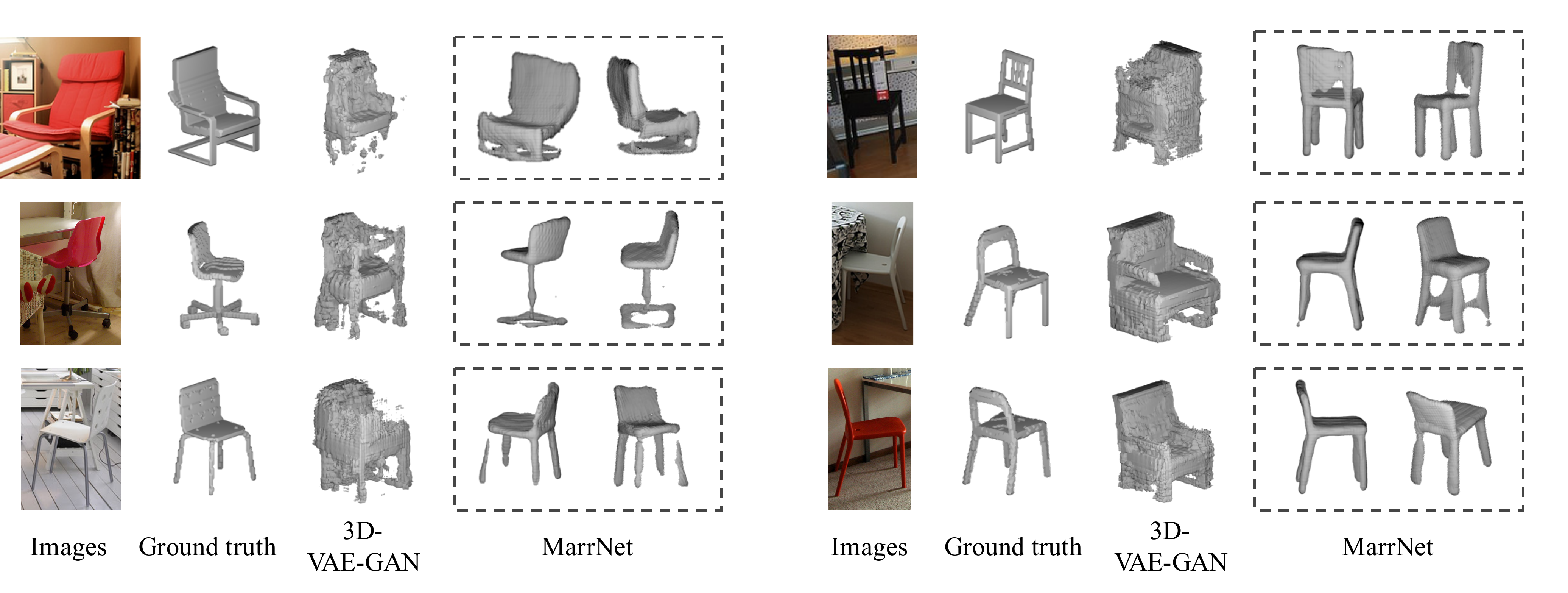}
    \caption{3D reconstruction of chairs on the IKEA~\citep{Lim2013} dataset. From left to right: input, ground truth, 3D estimation by 3D-VAE-GAN~\citep{Wu2016}, and two views of \model predictions. Our modelrecovers more details compared to 3D-VAE-GAN.}
    \label{fig:ikea}
\end{figure}

%% file: figText_multi.tex
\begin{figure}[t]
    \centering
    \includegraphics[width=\linewidth]{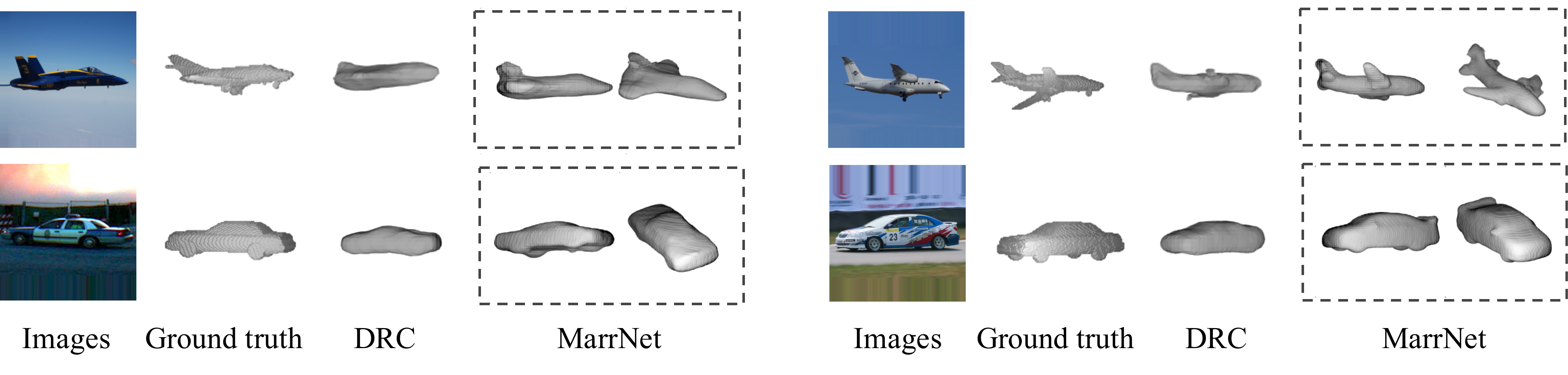}
    \caption{3D reconstructions of airplanes and cars from PASCAL 3D+. From left to right: input, the ground truth shape from the dataset, 3D estimation by DRC~\citep{tulsiani2017multi}, and two views of \model predictions.}
    \label{fig:multi}
\end{figure}

%% file: figText_multiclass.tex
\begin{figure}[t!]
    \centering
    \includegraphics[width=\linewidth]{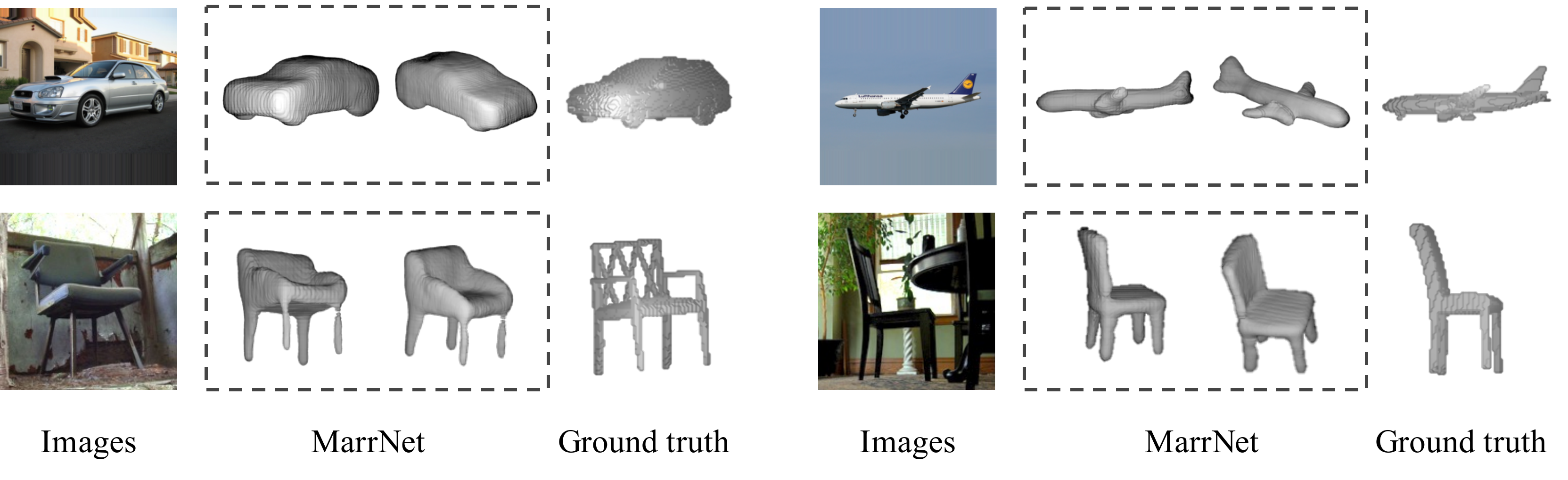}
    \caption{3D reconstruction of objects from multiple categories on the PASCAL 3D+~\citep{Xiang2014} dataset. Under this setup \model also recovers 3D shape well.}
    \label{fig:multiclass}
\end{figure}

%% file: discussion.tex
\section{Conclusion}

We proposed \model, a novel model that explicitly models 2.5D sketches for single image 3D shape reconstruction. The use of 2.5D sketches enhanced the model's performance, and made it easily adaptive to images across domains or even categories. We also developed differentiable loss functions for the consistency between 3D shape and 2.5D sketches, so that \model can be end-to-end fine-tuned on real images without annotations.  Experiments demonstrated that our model performs well, and is preferred by human annotators over competitors.